\documentclass{article}
\DeclareUnicodeCharacter{3001}{,}

\PassOptionsToPackage{numbers, compress}{natbib}

\usepackage[preprint]{neurips_2025}

\usepackage[utf8]{inputenc} 
\usepackage[T1]{fontenc}    
\usepackage{hyperref}       
\usepackage{url}            
\usepackage{booktabs}       
\usepackage{amsfonts}       
\usepackage{nicefrac}       
\usepackage{microtype}      
\usepackage{xcolor}         

\usepackage[pdftex]{graphicx} 
\graphicspath{ {./images/} }
\usepackage{multirow}  
\usepackage{subcaption}  
\usepackage{amsmath}
\usepackage{amssymb}  
\usepackage{float} 

\title{SCMPPI: Supervised Contrastive Multimodal Framework for Predicting Protein-Protein Interactions}

\author{
    Shengrui Xu\\
    Cuiying Honors College\\
    Lanzhou University\\
    \texttt{320220929071@lzu.edu.cn}
    \and
    Tianchi Lu\thanks{Corresponding author. Email: \texttt{tianchilu4-c@my.cityu.edu.hk}}\\
    Department of Computer Science\\
    City University of Hong Kong\\
    \texttt{tianchilu4-c@my.cityu.edu.hk}
    \and
    Zikun Wang\\
    School of Mathematics and Statistics\\
    Lanzhou University\\
    \texttt{320220936661@lzu.edu.cn}
    \and
    Jixiu Zhai\\
    School of Mathematics and Statistics\\
    Lanzhou University\\
    \texttt{320220948121@lzu.edu.cn}
}

\begin{document}

\maketitle

\begin{abstract}
Protein-protein interaction (PPI) prediction plays a pivotal role in deciphering cellular functions and disease mechanisms. To address the limitations of traditional experimental methods and existing computational approaches in cross-modal feature fusion and false-negative suppression, we propose SCMPPI—a novel supervised contrastive multimodal framework. By effectively integrating sequence-based features (AAC, DPC, ESMC-CKSAAP) with network topology (Node2Vec embeddings) and incorporating an enhanced contrastive learning strategy with negative sample filtering, SCMPPI achieves superior prediction performance. Extensive experiments on eight benchmark datasets demonstrate its state-of-the-art accuracy (98.13\%) and AUC (99.69\%), along with excellent cross-species generalization (AUC>99\%). Successful applications in CD9 networks, Wnt pathway analysis, and cancer-specific networks further highlight its potential for disease target discovery, establishing SCMPPI as a powerful tool for multimodal biological data analysis.
\end{abstract}

\section{Introduction}
Protein-Protein Interactions (PPI) are central to many biological processes within cells, including signal transduction, gene expression regulation, metabolic regulation, and the cell cycle \cite{alberts1998i1, chaplin2010i2, simons2000i3}. Accurately predicting PPIs is of great significance for understanding cellular functions, revealing disease mechanisms, and identifying potential drug targets \cite{marcotte1999d}. However, traditional experimental methods, such as yeast two-hybrid screening \cite{ito2001c1} and tandem affinity purification \cite{gavin2002C2}, are often time-consuming, labor-intensive, and costly. Therefore, in recent years, computational methods have become a popular research direction for PPI prediction, attracting considerable attention \cite{yao2019DEEPFE-PPI, chen2019PIPR, huang2020deeppurpose}.

With the rapid development of bioinformatics, multimodal models based on multiple data types, such as sequence, structural, and network information, have gradually become the mainstream approach for PPI prediction. These methods significantly improve prediction accuracy by integrating information from different sources. For example, DF-PPI \cite{DFPPI2024} enhances sequence feature capture and integrates handcrafted features and semantic embeddings, utilizing a dynamic weighted fusion strategy to improve model stability. TAGPPI\cite{TAGPPI2022}, on the other hand, combines AlphaFold-predicted structural features with sequence information and improves prediction accuracy using a graph neural network model.

Contrastive learning, as a effective self-supervised learning method, has gained increasing attention in the bioinformatics field in recent years \cite{chen2025TPpred-SC, zhang2024EPACT}, particularly for handling high-dimensional, unstructured biological data (protein sequences and structural information) and optimizing model generalization \cite{chen2020SIMCLR, khosla2020supCON}. However, to date, no studies have applied supervised contrastive learning to the PPI prediction task.

Existing multimodal PPI models still face several challenges: first, some methods rely on specific feature extraction methods, and the lack of high-precision protein structural data limits the application of these models. Second, the robustness and generalization of these models still need further improvement. Additionally, the problem of false-negative predictions persists.

To address these issues, this study proposes a novel supervised contrastive multimodal framework (SCMPPI), aimed at further improving PPI prediction performance by combining multimodal features with supervised contrastive learning. We perform sequence embeddings (AAC,DPC,ESMC-CKSAAP) and combine them with the Node2Vec graph embedding method to promote joint representation learning of sequence features and network information.For the PPI task, we enhance supervised contrastive learning by introducing a negative sample filtering mechanism and modifying the contrastive loss function. These improvements strengthen the model's ability to embed both sequence and network information, thereby reducing the false-negative rate. Experimental validation on multiple benchmark datasets shows that SCMPPI outperforms existing methods in terms of prediction accuracy, robustness, and generalization ability.

Our main contributions are as follows:
\begin{itemize}
    \item We propose a deep learning framework, SCMPPI, capable of integrating protein sequence features and ppi network information, combined with an improved supervised comparative learning strategy.
    
    \item SCMPPI not only achieves state-of-the-art performance on multiple benchmark datasets, but also exhibits good robustness, generalizability, and low false negatives.
    
    \item The multimodal collaborative mechanism achieved through contrastive learning provides a more versatile framework for interaction prediction, which is expected to advance the development of biomedical research.
\end{itemize}

\section{Related Works}
The related prior works include studies on multimodal models for PPI prediction and contrastive learning in biological tasks.
\paragraph{Multimodal Models for PPI}
In the field of PPI prediction, the exploration of multimodal models has driven the field toward greater technological complexity and data diversity. By integrating sequence, structural, and network information, these methods significantly enhance prediction accuracy. Models such as DF-PPI and TAGPPI have gained attention due to their advantages in combining multiple sources of information.
TAGPPI (Bosheng Song et al., 2022)\cite{TAGPPI2022} innovatively combines AlphaFold-predicted structural features with sequence information to construct a graph neural network model. However, its performance is limited by the accuracy of the structural predictions, and its robustness in low-quality data scenarios still requires further validation.
DF-PPI (Hoai-Nhan Tran et al., 2024)\cite{DFPPI2024}enhances sequence feature capture through an improved APAACplus descriptor and integrates handcrafted features with semantic embeddings using a dynamic weighted fusion strategy. It also improves model stability through batch normalization. However, its reliance on GPU resources and the redundancy of handcrafted features may limit its large-scale application.

In this study, we use the ESMC protein language model\cite{esmc2024cambrian} to extract sequence embeddings, combining them with Node2Vec graph embedding methods\cite{node2vec} to achieve joint representation learning of sequence features and network information.
In parallel with research on multimodal methods in the bioinformatics PPI field, the superiority of contrastive learning has been recognized, and attempts are being made to apply contrastive learning to biological information tasks.
\paragraph{Supervised Contrastive Learning in Biological Tasks}
In the field of bioinformatics, supervised contrastive learning has emerged as a effective feature embedding method, demonstrating its unique advantages in various tasks. By optimizing the feature distributions between positive and negative samples while incorporating positive sample label information, contrastive learning enhances the model's ability to distinguish between similar and dissimilar samples. This makes it particularly suitable for handling high-dimensional, non-structured biological data such as protein sequences and structural information. Consequently, contrastive learning has been widely applied to improve models' ability to embed sequence and structural information. For instance, the EPACT (Yumeng Zhang et al., 2024) \cite{zhang2024EPACT}framework harnesses contrastive learning to combine with the pre-trained protein language model (TAPE), significantly improving the prediction of T cell receptor-antigen binding specificity. However, the framework's use of negative sampling strategies may introduce unavoidable biases, leading to higher false negative probabilities [42]. Additionally, TPpred-SC (Ke Yan et al., 2023)\cite{chen2025TPpred-SC} extends contrastive learning to multilabel classification tasks, predicting peptides with multiple functional attributes. These methods not only enhance prediction accuracy but also improve the model's ability to handle long sequences and complex networks.

In this study, we apply supervised contrastive learning \cite{khosla2020supCON}to the PPI task. By pulling positive sample pairs closer, pushing negative sample separation farther apart,and extracting label information, we have improved the ability of the multimodal model to embed sequence and structural information. We also incorporate a negative sample filtering mechanism to reduce the likelihood of false negatives and modify the loss of SuoCons $\mathcal{L}^{Sup}_{out}$ to $\mathcal{L}^{P-Sup}$ for PPI.

\section{Approach}
\label{approach}


This section delineates the architecture of the SCMPPI framework, an innovative concept devised to facilitate efficient and high-quality PPI prediction.The structural and design principles of the framework are meticulously expounded in Section 3.1, followed by an overview of the application of multimodal fusion techniques in Section 3.2, and finally its core contrast learning module (Section 3.3) and classifier (Section 3.4), respectively.


\subsection{Architecture of SCMPPI}
\begin{figure}[h]
    \centering
    \includegraphics[width=1\linewidth]{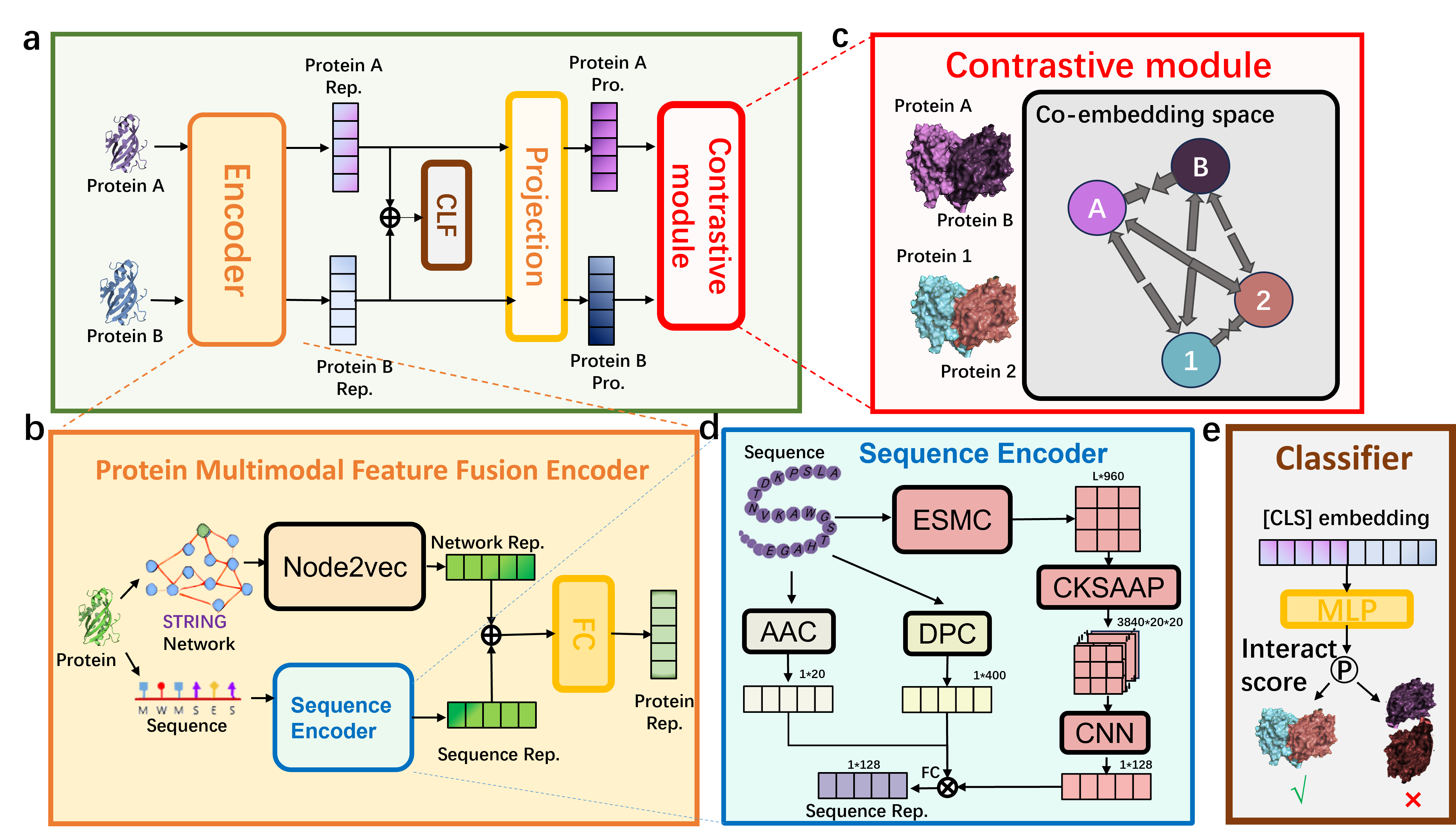}
    \caption{The Architecture of SCMPPI.(a) SCMPPI includes a protein encoder, interaction classifier, and contrastive learning module.(b) Encoder has sequence encoder and Node2vec.(c) Contrastive module projects embeddings into shared space and calculates loss using cosine distance.(d) Sequence encoder uses AAC, DPC, and ESMC-CKSAAP for features.(e) Predicts interaction by outputting a score.}
    \label{fig:Architecture of SCMPPI}
\end{figure}

We employed a divide-and-conquer paradigm to develop the architecture of SCMPPI.The detailed model architecture is presented in  \autoref{fig:Architecture of SCMPPI}. The protein multimodal feature fusion encoder primarily consists of a sequence encoder and Node2vec, which respectively extract sequence features of the protein and embedding features from the protein interaction graph. These two features are fused through a MLP to form the representation of the protein.The  embeddings of protein pairs are then fused to provide model predictions for various downstream tasks or are fed into the contrastive co-embedding module.In summary, the protein sequences are sampled and fed into the multimodal protein encoder to obtain the embeddings for the proteins. These effective embeddings are then applied to downstream tasks for various model predictions.

\subsection{Multimodal Fusion Encoder}
Overall, the encoder extracts protein interaction graph information through Node2vec and protein sequence information through the Sequence Encoder, and then integrates them into a fixed-length protein representation using FC.

\paragraph{Node2vec}
We use the NW-align tool \cite{kaur2017NW-ALIGN} to retrieve similar proteins from the STRING database \cite{mering2003string}, finding an interaction network of similar proteins. Then, we remove edges related to the original protein to avoid data leakage \cite{avoiding_data_leakage}. Next, Node2vec captures the graph embeddings of these similar proteins, which are used as the original protein's graph embeddings.

In the PPI task, we focus on two types of similarity: vertex homogeneity and structural equivalence \cite{N1,N2,N3}. Vertex homogeneity suggests that proteins with similar features are more likely to interact, while structural equivalence indicates that proteins with similar roles in the network tend to interact. These similarities help predict protein interactions \cite{N4}.

To capture these attributes, we generate the protein node context using biased random walks. The transition probability from protein node $t$ to protein node $x$ is defined as:
\[
\pi_{tx} = \alpha_{pq}(t, x) \cdot w_{tx}
,\tag{1}\]
where $w_{tx}$ is the edge weight and $\alpha_{pq}(t, x)$ is the search bias. This allows Node2vec to explore both local and global graph structures and learn meaningful protein node embeddings.

\paragraph{Sequence Encoder}
The proposed sequence encoder transforms protein sequences into high-dimensional vector representations for downstream tasks like function prediction, classification, or structural analysis. It integrates three feature extraction strategies: AAC, DPC, and ESMC-CKSAAP, capturing a range of sequence features, from global composition to local structural patterns and long-range dependencies.

\textit{(i) AAC (Amino Acid Composition)} \cite{aac_dpc_prediction} calculates the frequency of each amino acid in the sequence, producing a 20-dimensional vector:
\[
\text{AAC}(aa_i) = \frac{\text{count of amino acid } aa_i}{L},\tag{2}
\]
where $L$ is the sequence length. This representation highlights the overall distribution of amino acids.

\textit{(ii) DPC (Dipeptide Composition)} \cite{aac_dpc_prediction} extends this by considering adjacent amino acid pairs, resulting in a 400-dimensional vector. The formula is:
\[
\text{DPC}(aa_i, aa_{i+1}) = \frac{\text{count of dipeptide } (aa_i ,aa_{i+1})}{L - 1},\tag{3}
\]
This captures short-range interactions within the protein sequence.

\textit{(iii) CKSAAP (k-gap amino acid pairs)} \cite{chen2011CKSAAP} protein input to the ESMC model produces amino acid-level embeddings \((ae_1, ae_2, \dots, ae_L)\) with a shape of [L, 960]. The mean of amino acid pair embeddings replaces frequency in the CKSAAP formula to produce:
\[
\text{ESMC-CKSAAP}(aa_i, aa_{i+1}) = \frac{\text{mean of }(ae_i, ae_j) \text{ pairs at distance } k}{\text{sum embeddings at distance } k},\tag{4}
\label{eq:k}
\]
This produces an embedding of size [20, 20, 960], and traversing the distance \(k\) from 0 to 3 results in a tensor with a shape of \([k+1, 20, 20, 960]\). This tensor can be reshaped based on downstream tasks, adjusting to a 3D shape \([k \times 960, 20, 20]\) for 3D convolution operations. In our work, with \(k=3\), the final reshaped embeddings have a shape of [3840, 20, 20].For parameter details, refer to \autoref{app:para}.This integration utilizes the biophysical and evolutionary insights of the ESMC model, embedding the contextual information of amino acid pairs up to a gap of \(k\), reflecting residue correlations and highlighting local interactions within the protein.

AAC captures the overall amino acid composition, DPC focuses on local structure through adjacent pairs, and CKSAAP with ESMC embeddings captures long-range dependencies and contextual features. This multi-scale fusion provides a rich, comprehensive sequence representation, ideal for bioinformatics tasks.

\subsection{Contrastive Learning Module}
\paragraph{Contrastive Learning Loss}
Contrasive learning Loss of SCMPPI is based on the unsupervised contrastive learning method SimCLR \cite{chen2020SIMCLR}. The loss of SimCLR can be formulated as :
\begin{equation} 
\mathcal{L}^{Sim} = \sum_{i \in I} \mathcal{L}^{Sim}_i = - \sum_{i \in I} \log \frac{\exp(z_i \cdot z_{j(i)}/\tau)}{\sum_{a \in A(i)} \exp(z_i \cdot z_a/\tau)}, \tag{5}
\end{equation} 
where $i \in I = {1, 2, \dots, 2N}$ represents index of augmented samples in training batches of size $N$. While $j(i)$ indicates the index of another augmented sample from the same source sample. $A(i) = I \setminus {i}$ represents the set of indices excluding $i$. $\tau$ represents the temperature parameter, which controls the penalty strength for hard negative samples \cite{wang2021TEMP}. Specifically, smaller temperature values lead to stronger penalties for the most difficult negative samples, representing a greater similarity between vectors. The sample indexed by $j(i)$ is the positive sample related to $i$, while all other samples are considered negative samples. In SimCLR, each anchor sample has only one positive sample, resulting in $2N - 1$ negative samples.

While SimCLR is typically used for pre-training on large unlabeled datasets, it ignores label information. SupCons \cite{khosla2020supCON} proposed a method to incorporate labels into the loss function, enabling effective utilization of labeled information. Its formula is formulated as:
\begin{equation} \mathcal{L}^{Sup}_{out} = \sum_{i \in I} \mathcal{L}^{Sup}_i = - \frac{1}{|P(i)|} \sum_{p \in P(i)} \log \frac{\exp(z_i \cdot z_p/\tau)}{\sum_{a \in A(i)} \exp(z_i \cdot z_a/\tau)}, \tag{6} \end{equation} 
where  $P(i) = \{p| p \in A(i) \land y_p = y_i\}$  represents the set of positive samples, which are samples with the same label as $i$. Compared to SimCLR, SupCons expands the positive sample set for each anchor sample, effectively utilizing label information from positive samples. However, SupCons only utilizes positive sample label information and does not effectively utilize negative sample label information, assuming that all non-anchor samples of different classes are negative samples with low false negative probabilities.

To better collect negative sample information and reduce the false negative probability in this work, we modified SupCons to obtain the protein supervised contrastive learning loss ($\mathcal{L}^{p-Sup}$):
\begin{equation} 
\mathcal{L}^{P-Sup} = \sum_{i \in I} \mathcal{L}^{P-Sup}_i = - \frac{1}{P(i)} \sum_{p \in P(i)} \log \frac{\exp(z_i \cdot z_p/\tau)}{\sum_{a \in A(i)} \exp(z_i \cdot z_a/\tau) \cdot I_{\{t_{ia} \leq \tau \} }(x)}, \tag{7} 
\label{eq:t}
\end{equation} 
\begin{equation} 
t_{iq} =\frac{\sum_{j} z_{i,j} \cdot z_{q,j}}{\sqrt{\sum_{j} z_{i,j}^2} \cdot \sqrt{\sum_{j} z_{q,j}^2} + \epsilon},
\tag{8} \end{equation}
where $t_{iq}$ represents the Cos-score \cite{azizian2024similar} between sample $i$ and sample $q$, used to measure the similarity between samples $i$ and $q$. The term $I_{\{ t_{ia} \leq \tau \} }(x)$ filters negative samples to reduce the false negative probability, where samples $i$ and $q$ with a Cos-score similarity score below the threshold $\tau$ are accepted as negative samples.

When minimizing $\mathcal{L}^{p-Sup}$, samples with higher similarity scores are clustered more tightly in the feature space. Conversely, they are drawn apart.
\paragraph{Co-embedding Space}
The two protein representations generated by the Multimodal Fusion Encoder are projected into a shared latent space using the Projection projector. In our contrastive learning approach, one protein is designated as the anchor, and the associated binding proteins are pulled closer to the anchor in the latent space, while "non-binding" (negative) proteins are pushed away. Given a protein, a set of binding (positive) proteins, and a set of decoy (negative) proteins along with their projections in the training batch, the cosine similarity between protein pairs is calculated.
\begin{figure}[ht]
\centering
\begin{minipage}[b]{0.45\textwidth}
    \centering
    \includegraphics[width=\textwidth]{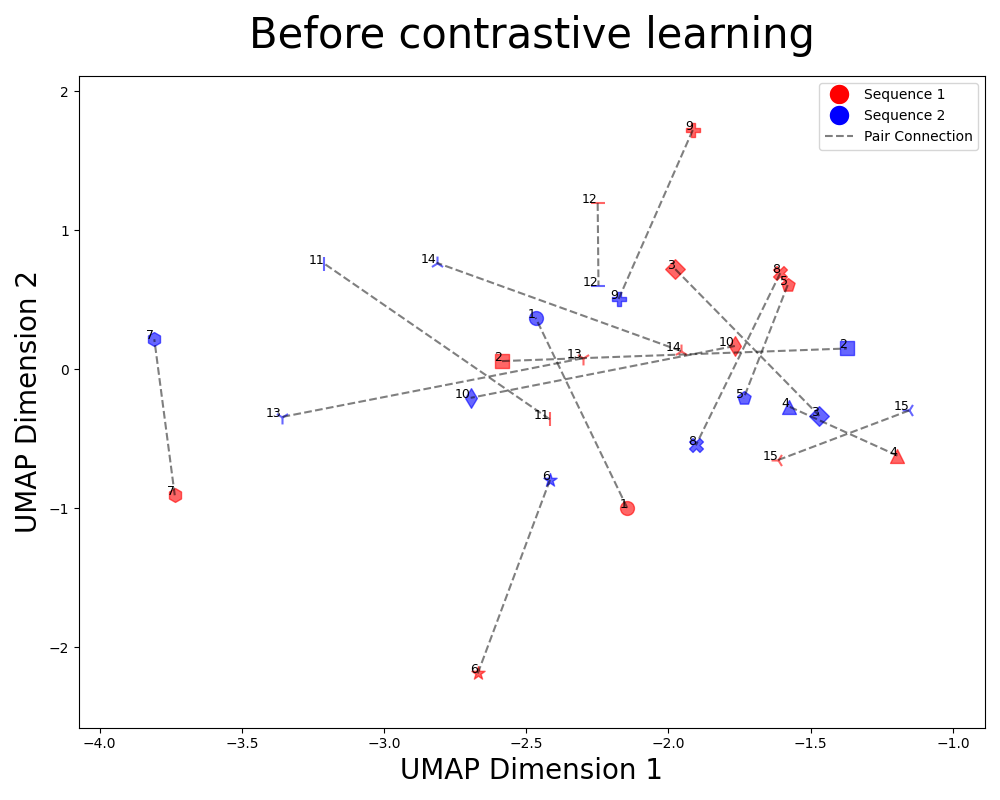}  
\end{minipage}%
\hspace{0.05\textwidth} 
\begin{minipage}[b]{0.45\textwidth}
    \centering
    \includegraphics[width=\textwidth]{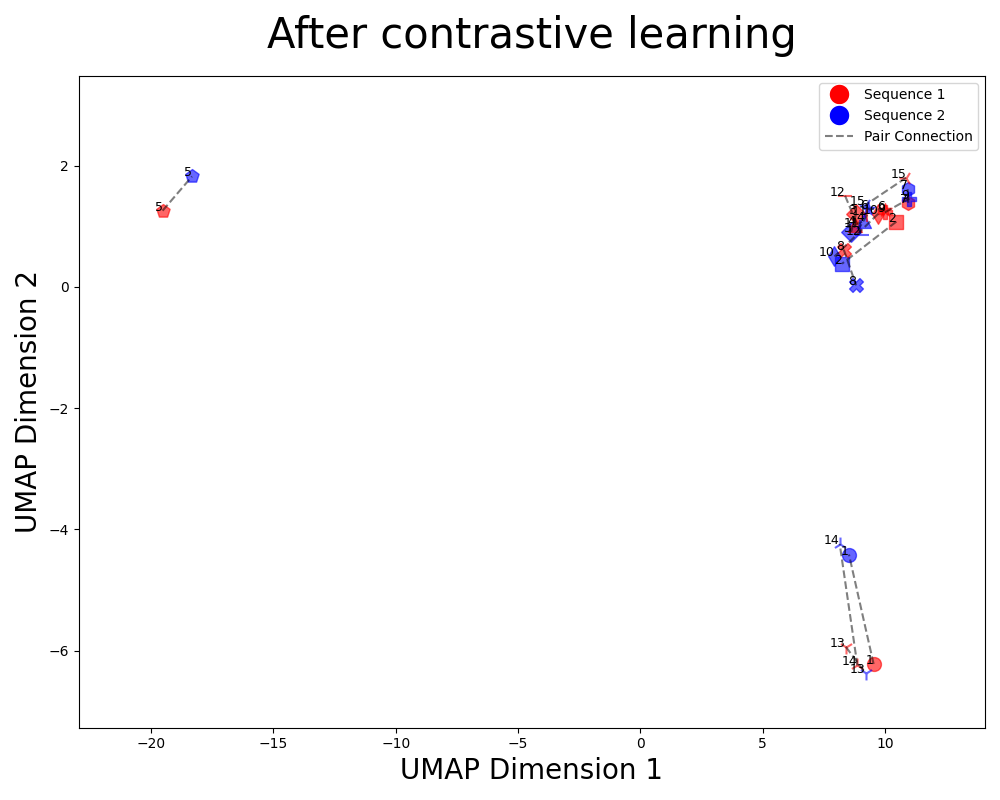}  
\end{minipage}
\caption{Comparison of results before and after contrastive learning}
\label{fig:cl}
\end{figure}
We use the Uniform Manifold Approximation and Projection(UMAP)\cite{mcinnes2018umap} to reduce the dimensionality of the projections, which results in different protein embeddings in a two-dimensional space. In \autoref{fig:cl}, we randomly selected 15 pairs of binding (positive) proteins from the independent test set and projected their corresponding embeddings onto 2D UMAP plots, both before and after contrastive learning. A comparison of the two plots shows that the embeddings are closer together after contrastive learning, demonstrating the effectiveness of contrastive learning in PPI.

\subsection{Prediction Module}
In a training batch of $N$ protein sequence pairs $(S_i, S_j)$, each pair generates latent vectors $\mathbf{F_i}$ and $\mathbf{F_j}$ through
 Protein Multimodal Feature Fusion Encoder. By concatenating them, we obtain the classification embedding $\mathbf{F_{ij}} = [\mathbf{F_i}; \mathbf{F_j}]$, which is passed through an MLP to produce the classification result:
\[
\hat{y}_{ij} = \text{MLP}(\mathbf{F_{ij}}),\tag{10}
\]
where $\hat{y}_{ij}$ is the predicted interaction probability between $(S_i, S_j)$.

The binary cross-entropy loss for a batch of $N$ inputs $(S_i, S_j, y_{ij})$ is:
\[
\mathcal{L}_{BCE} = - \sum_{i=1}^{N} \sum_{j=1}^{N} \left( y_{ij} \log(\hat{y}_{ij}) + (1 - y_{ij}) \log(1 - \hat{y}_{ij}) \right),\tag{11}
\]
and the total loss function is:
\[
\mathcal{L} = \mathcal{L}_{BCE} + \kappa \mathcal{L}^{p-\text{Sup}},\tag{12}
\]
where $\kappa$ controls the balance between contrastive loss $\mathcal{L}^{p-\text{Sup}}$ and classification loss $\mathcal{L}_{BCE}$, optimizing the model’s generalization and robustness. For parameter details, refer to \autoref{app:para}.

\section{Experiments}
\label{experiments}
In this section, we describe the experimental setup we used and analyse what we found.

\subsection{Experimental Settings}

\subsubsection{Baseline Algorithms}
In our experiments, we represent our algorithm as SCMPPI and compare it with traditional optimizers and existing robust methods for PPI prediction. The traditional optimizers considered are Decision Tree (DT) and Random Forest (RF). The existing robust methods for PPI prediction include DeepFE-PPI\cite{yao2019DEEPFE-PPI}, PIPR\cite{chen2019PIPR}, KSGPPI\cite{hu2024KSGPPI}, 
TAGPPI\cite{TAGPPI2022}, 
DF-PPI\cite{DFPPI2024}, 
DeepTrio\cite{hu2022deeptrio}, 
DeeP-AAC\cite{huang2020deeppurpose}, 
DeeP-CNN\cite{huang2020deeppurpose}, DFC\cite{srivastava2012DFC,xie2023hnsppi}, DCONV\cite{yang2020graph,xie2023hnsppi}, and HNSPPI\cite{xie2023hnsppi}. In this experiment, we utilize a five-fold cross-validation method, which has been widely adopted in previous studies\cite{chen20225f1,gao20235f2,chen20195f3}. Additionally, as mentioned earlier, the best configurations of the above models were established as described in their respective works.

\subsubsection{Datasets}
We conducted experiments using eight benchmark PPI datasets. The Yeast PPI dataset\cite{hashemifar2018yeast1,wong2015yeast2,you2014yeast3} contains 2,497 proteins and 11,188 protein-protein interactions (PPIs), with a balanced sample distribution and complete sequences. 
The Human PPI dataset, constructed by Pan et al.\cite{pan2010human}, includes 3,899 positive samples and 949 negative samples, representing 2,502 proteins. To balance the samples, 2,950 new negative protein pairs were generated using the method.\cite{xie2023hnsppi} The H.pylori dataset was originally derived from Rain's work \cite{rain2001pylori} and consists of 1458 protein pairs involving 1313 different proteins.
The Yeast (PIPR-cut) dataset\cite{hu2024KSGPPI} is derived from the Yeast (PIPR) dataset, comprising 4,487 positive samples and 4,487 negative samples, totaling 8,974 PPIs and involving 2,039 proteins.
The multi-species dataset\cite{chen2019multi} covers Escherichia coli, Caenorhabditis elegans, and Drosophila melanogaster, with 1,834, 2,637, and 7,058 proteins, respectively. The sample sizes are 6,954, 4,013, and 21,975 negative samples, each with an equal number of positive pairs. 
Finally, the CD9 network, Wnt-related pathway, and Cancer-specific network datasets consist of 16, 96, and 108 samples, respectively\cite{DFPPI2024}. 
These datasets provide a rich and diverse set of resources for model evaluation. For detailed datasets information, refer to \autoref{app:dateset}

\subsection{Results Analysis}

\subsubsection{Performance of SCMPPI on intraspecies dataset}
\begin{table}[h!]
\centering
\caption{Five-fold cross-validation results of the SCMPPI on the intraspecies dataset}
\label{tab:int}
\vspace{0.7em}
\begin{tabular}{l l l l l l l}
\toprule
  & Acc(\%) & Pre(\%) & Sen(\%) & F1(\%) & AUC(\%) & AUPRC(\%) \\
\midrule
Yeast & 98.13 & 97.75& 98.32& 98.17& 99.58& 99.53\\
Human & 83.02 & 81.84 & 84.95 & 83.34 & 90.58 & 90.84 \\
H.pylori & 87.67 & 84.12 & 93.28 & 88.45 & 93.99 & 93.13 \\
PIPR-cut & 88.54 & 88.12 & 89.12 & 88.61 & 94.46 & 92.42 \\
\bottomrule
\end{tabular}
\end{table}

\autoref{tab:int} shows the 5-fold cross-validation results of the SCMPPI model on the intraspecies datasets. As seen in the table, the SCMPPI model performs excellently and stably across different datasets. The accuracy (acc) and sensitivity (sen) evaluated on the Yeast dataset are both above 98\%. AUC and AUPRC represent the model's ability to distinguish between positive and negative samples at different thresholds and its performance evaluation when handling imbalanced data. The AUC and AUPRC values for SCMPPI on the four intraspecies datasets are all above 90\%, demonstrating that our model has high discriminative power and the ability to handle imbalanced data across various datasets, providing a reliable solution for  PPI prediction.

\subsubsection{Comparisons with existing algorithms}
\begin{table}[h!]
\centering
\caption{Performance comparison on dataset Yeast ($\kappa$=1.0).}
\label{tab:yeast}
\vspace{0.7em}  
\begin{tabular}{l l l l l l l l}
\toprule
Model  & Acc(\%)  & Pre(\%)  & Sen(\%)  & F1(\%)  & MCC  & AUC(\%)  & AUPRC(\%)  \\
\midrule
SCMPPI & \textbf{98.13}& 97.75& \textbf{98.53}& \textbf{98.14}& \textbf{0.962}& \textbf{99.69}& \textbf{99.69}\\
DF-PPI & 96.34 & 97.56 & 95.05 & 96.29 & 0.927 & 98.87 & 99.16 \\
TAGPPI & 97.81 & \textbf{98.10} & 98.26 & 97.80 & 0.956 & 97.74 & - \\
KSGPPI & 97.64 & 97.44 & 97.85 & 97.62 & 0.956 & 97.25 & 97.99 \\
PIPR & 97.09 & 97.00 & 97.17 & 97.09 & 0.942 & - & - \\
DeepFE-PPI & 94.78 & 96.45 & 92.99 & 94.69 & 0.896 & 98.83 & 98.53 \\
RF & 93.62 & 96.75 & 90.26 & 93.40 & 0.874 & 96.52 & 97.27 \\
DF & 87.78 & 88.47 & 86.86 & 87.65 & 0.756 & 87.78 & 83.42 \\
\bottomrule
\end{tabular}
\end{table}

To validate the superior performance, we compared the proposed model, SCMPPI, with existing robust methods for PPI prediction on all four species-specific benchmark datasets. As shown in Tables \ref{tab:yeast}, \ref{tab:human}, \ref{tab:pylori}, and \ref{tab:pipr}, SCMPPI delivers the best performance across the Yeast, Human, H. pylori, and PIPR-cut datasets.

The Yeast dataset is a widely recognized benchmark in the field of PPI and is commonly used to evaluate the performance of advanced methods\cite{hashemifar2018yeast1,wong2015yeast2,you2014yeast3}. We compared our SCMPPI model with the following representative methods on the Yeast dataset: DE-PPI, TAGPPI, KSGPPI, PIPR, DeepFE-PPI , RF, and DF. To ensure a fair comparison, we maintained the structural parameters and hyperparameters of these methods. The performance of these methods was evaluated using five-fold cross-validation on different datasets .
As seen in Table \ref{tab:yeast} and Figure \ref{fig:box}, on the Yeast dataset, SCMPPI achieved the best performance across almost all evaluation metrics. By integrating sequence and network features using a multimodal encoder and learning the latent space distances through contrastive learning, SCMPPI improved accuracy, sensitivity (Sen), and Matthews correlation coefficient (MCC) by 3.48\%, 5.90\%, and 7.37\%, respectively, compared to DeepFE-PPI. Compared to KSGPPI, our model improved AUC by 2.49\% and AUPRC by 1.59\%.
Notably, SCMPPI achieved the highest sensitivity (Sen) of 98.48\% and F1 score of 98.09\% among all the methods. F1 score and sensitivity are crucial metrics for evaluating a model's ability to identify positive instances. A high sensitivity indicates that the model can recognize most of the positive samples. This is particularly important as SCMPPI is less likely to incorrectly predict interacting protein pairs as non-interacting, resulting in fewer false negatives (FN) compared to other methods.
For Precision (Pre) , our model ranked second compared to the other methods, slightly lower than the best TAGPPI by 0.34\%. However, on harmonic measurements such as AUC and MCC, which are essential for a binary classifier, our model outperformed TAGPPI by 1.97\% and 0.63\%, respectively. This demonstrates that our model strikes a harmonious balance across all evaluation metrics, achieving high accuracy and reliability.

In evaluating model performance, the stability of predictions is also an important factor. As shown in Figure \ref{fig:box}, our model demonstrates high stability across all evaluation metrics. Specifically, the DT model shows relatively long boxes and whiskers in Accuracy (Acc), Sensitivity (Sen), AUC, F1, AUPRC, and MCC, indicating considerable fluctuations in the metrics and poor stability. In addition to the traditional models, DeepFE-PPI also exhibits poor stability in Precision (Pre), Specificity (Spec), F1, and MCC. These findings highlight the reliability of our fusion feature strategy in PPI prediction.

\subsubsection{Robustness and Generalization Study}

\begin{table}[b]
\centering
\caption{Train on Yeast dataset and test on H.pylori or Human dataset}
\label{tab:test}
\vspace{0.7em}  
\begin{tabular}{l l l l l ll l }
\toprule
Test dataset & Model & Acc\% & Pre\% & F1\% &  MCC 
&AUC\% & AUPRC\% \\
\midrule
\multirow{3}{*}{H.pylori} 
& SCMPPI & \textbf{58.41} & \textbf{59.04} & \textbf{58.74} &  \textbf{0.168} 
&\textbf{60.04} & \textbf{57.04} \\
& KSGPPI & 56.60 & 58.20 & 54.30 &  0.134 
&59.32 & 56.63 \\
& DeepPE-PPI & 48.40 & 47.71 & 27.45 &  -0.030 
&48.08 & 48.74 \\
\midrule
\multirow{3}{*}{Human} 
& SCMPPI & \textbf{55.26} & \textbf{55.36} & \textbf{54.82} &  \textbf{0.1052} 
&\textbf{57.23} & \textbf{55.79} \\
& KSGPPI & 53.59 & 52.41 & 62.70 &  0.0823 
&55.57 & 55.01 \\
& DeepPE-PPI & 49.04 & 46.11 & 18.30 &  -0.0292 &46.98 & 47.66 \\
\bottomrule
\end{tabular}
\end{table}

For PPI prediction, robustness and generalization are crucial because the model needs to accurately predict PPIs across different datasets and data distributions. Therefore, we evaluated the robustness and generalization of our model, and the results show that SCMPPI performs exceptionally well in both aspects.

\autoref{tab:multi-species} shows the performance of several models on the multi-species dataset (including C. elegans, D. melanogaster, and E. coli). SCMPPI outperforms other comparative models in multiple metrics, including accuracy, precision, and F1 score, demonstrating its stronger adaptability and robustness in predicting PPIs across different species. The model effectively overcomes data distribution differences, showing greater resistance to data noise and outliers while maintaining more stable prediction performance.

Regarding generalization, SCMPPI performs excellently, especially in \autoref{tab:test}, where the Yeast dataset is used for training and tested on H.pylori and Human datasets, it still maintains the highest prediction performance. This indicates that SCMPPI not only achieves excellent results on specific datasets but also effectively transfers and adapts to new data distributions across datasets, further validating its wide applicability and strong generalization ability.

\subsubsection{Ablation Study}
Through ablation experiments, we assessed the contributions of the sequence module, graph module, and contrastive learning module in PPI prediction.
The results from ablation experiments conducted on multiple datasets (Yeast, Human, H. pylori, and PIPR-cut), as shown in \autoref{tab:xiaorong_comparison}, indicate that the combination of the sequence (Seq), graph (Graph), and contrastive learning (Cl) modules significantly enhances the model's performance. In all of the datasets that were analysed, the model demonstrated the greatest performance when employing a combination of the three modules (Seq, Graph and Cl). The model demonstrated quasi-optimality in a range of metrics, including accuracy (Acc), the Matthews correlation coefficient (MCC), sensitivity (Sen), the area under the curve (AUC), and the area under the precision-recall curve (AUPRC).
When any of the modules (Seq, Graph, or Cl) is removed, the performance drops significantly, highlighting the indispensable contribution of each module to the overall prediction capability.

Overall, the ablation experiments validate the complementary nature of the three modules, demonstrating that they not only improve the model's robustness, generalization ability, and accuracy but also ensure the efficiency and accuracy of the PPI prediction task.

\subsubsection{Testing on PPI Network Datasets}
\label{sec:net}

In this study, we applied the SCMPPI model to predict PPIs within three key biological networks : CD9, Wnt-related pathway, and Cancer-specific networks. The model successfully identified critical PPIs in the \autoref{fig:net}, revealing biologically and clinically relevant proteins \cite{rapposelli2021netfor1,dimitrakopoulos2023netfot2}.





Firstly, CD9 is a tetraspanin transmembrane protein widely distributed across various cell types, involved in processes such as cell adhesion, migration, and fusion source\cite{jiang2024cd9}. In tumor metastasis, CD9 plays a dual role. In some cancers, such as prostate cancer, high expression of CD9 is associated with lower metastatic ability, exhibiting a tumor-suppressive effect. However, in other types of cancer, such as pancreatic cancer, high CD9 expression correlates with poor prognosis. Therefore, the role of CD9 in tumor prognosis and treatment should be analyzed based on specific cancer types.

The Wnt-related pathway plays a crucial role in embryonic development, cell proliferation, differentiation, migration, and tissue homeostasis. Abnormal activation of this pathway is closely associated with the onset and progression of various cancers source\cite{song2024wnt}. AXIN1 and WNT9A are core proteins in this pathway. AXIN1 inhibits the activity of the Wnt pathway by forming a complex that degrades $\beta$-catenin, while WNT9A, as a ligand protein, activates the pathway. Regulating the activity of the Wnt signaling pathway holds potential clinical value for cancer prevention and treatment.

Finally, in the Cancer-specific network, CDK1 and TP53 are two key proteins. CDK1 regulates cell proliferation during the G2/M phase of the cell cycle, and its abnormal activation can promote tumor cell proliferation source\cite{malumbres2014cancer1}. TP53, as a tumor suppressor gene, is responsible for DNA repair and apoptosis, and mutations in TP53 are commonly found in various cancers source\cite{aubrey2016cancer2}. Studying these proteins and their interactions can provide an important theoretical foundation for cancer therapy.

Through the accurate predictions of the SCMPPI model, we have not only validated the important roles of these core proteins in various biological processes but also opened up new research directions for the early diagnosis and targeted treatment of cancer.

\section{Limitation and Conclusions}
\label{sec:lim and con}
In this study, we introduced the SCMPPI framework, which offers an innovative solution for protein-protein interaction (PPI) prediction by combining multimodal features with supervised contrastive learning. The multimodal collaborative mechanism achieved through contrastive learning provides a more versatile solution for interaction analysis. The method uses a supervised contrastive learning objective to align sequence semantics (based on ESMC embeddings) and structural topology (from known graphs), effectively advancing multimodal representation learning and significantly improving feature alignment integrity and robustness.
Experimental results show that SCMPPI outperforms existing methods in terms of both accuracy and generalization across multiple benchmark datasets, and it provides reliable results in PPI network prediction.

However, we acknowledge certain limitations. Despite the superior performance of SCMPPI, its complexity and computational demands still need to be reduced, especially when handling large-scale datasets. Furthermore, the model relies on specific feature extraction methods, such as Node2Vec (for graph embeddings) and sequence-based embeddings, which may limit its adaptability in some biological environments.
Future work could explore multiscale graph neural network architectures and incorporate dynamic negative sample selection strategies to optimize feature space distribution.


\bibliographystyle{unsrt}
\bibliography{reference}
\newpage
\appendix
\section{Detailed Datasets and Parameter settings}
\label{sec:d and p}
\subsection{Detailed Datasets}
\label{app:dateset}
Baseline data are broadly categorized into two types of datasets: intraspecific (Yeast, Human, H. pylori, PIPR-cut) and interspecific datasets (multi-species).

\textbf{Yeast}.The Yeast PPI dataset is a widely recognized benchmark used to evaluate the performance of advanced methods. This dataset contains 2,497 proteins, resulting in a total of 11,188 PPI pairs, split evenly into positive and negative samples. The positive samples are primarily derived from the DIP database, while negative interactions are generated by randomly pairing proteins without supporting evidence of interaction. The protein sequences in the dataset are complete and sourced from UniProt. To ensure dataset quality, protein sequences with fewer than 50 amino acids or sequences with 40\% or more identity were removed using CD-HIT. In the end, the dataset includes 5,594 positive protein interaction pairs and 5,594 negative samples, ensuring balance between positive and negative samples. This high-quality PPI data provides a reliable benchmark for related research.

\textbf{Human}.The Human PPI dataset was created by Pan et al., contains 3,899 positive samples and 949 negative samples, involving 2,502 human proteins. To balance positive and negative samples, 2,950 negative protein pairs were generated using the same method as the one used in the ksgppi dataset.

\textbf{H.pylori}.The H.pylori dataset, initially provided by Rain et al., includes 1,549 proteins from Helicobacter pylori, with 1,458 positive samples and 1,390 negative samples, providing a rich data foundation for related research.

\textbf{PIPR-cut}.The Yeast (PIPR-cut) dataset is derived from the Yeast (PIPR) dataset. To address redundancy within interaction pairs, repeated positive samples and those with an NW-alignment score above 0.4 were removed. After filtering, the final dataset contains 4,487 positive samples involving 2,039 proteins. To balance the number of positive and negative samples, negative samples were generated based on the positive proteins. A total of 2,966 negative samples from the PIPR dataset were retained, with the remaining 1,521 negative samples generated using the same method as the Human dataset. The final dataset includes 8,974 samples and 2,039 proteins.

\textbf{multi-species}.For evaluating the model's performance in cross-species PPI prediction, we used a multi-species dataset. This dataset combines multiple benchmark datasets to test the model's ability to predict protein interactions with low sequence identity across species.  
The specific datasets used in this study include Escherichia coli (E. coli), which consists of 1,834 proteins, 6,954 positive samples, and 6,954 negative samples; Caenorhabditis elegans (C. elegans), containing 2,637 proteins, 4,013 positive samples, and 4,013 negative samples; and Drosophila melanogaster (D. melanogaster), which includes 7,058 proteins, with 21,975 positive samples and 21,975 negative samples.

\textbf{network}. The three PPI network datasets used in our study include the CD9 network, which consists of 16 samples, focusing on specific protein core modules or functional units; the Wnt-related pathway, containing 96 samples, which focuses on proteins involved in the Wnt signaling pathway; and the cancer-specific network, with 108 samples, specifically constructed to explore cancer-related protein interactions.
These datasets represent different biological contexts and provide valuable resources for studying protein interactions in various biological processes. The CD9 network dataset focuses on a specific protein core module or functional unit, the Wnt-related pathway  investigates proteins involved in Wnt signaling, and the Cancer-specific targets cancer-related protein interactions. 

These datasets contribute to a deeper understanding of protein functions and interactions in biological processes and offer important data for disease diagnosis and therapeutic target discovery.

\subsection{Parameter settings}
\label{app:para}

\subsubsection{Evaluation Configuration}
\label{app:config}
All our experiments were conducted on an NVIDIA RTX 4070 GPU,an AMD Ryzen 9 7945HX with Radeon Graphics 2.501 GHz CPU, using Python 3.12.8 and PyTorch 2.6.0.

\subsubsection{Model Settings}\
The model utilizes convolutional layers (two Conv2D layers and a pooling layer) for feature extraction, followed by fully connected layers for sequence representation. It is trained using the AdamW optimizer\cite{kingma2014adam} with a learning rate of 0.001 and a batch size of 32. 
Early stopping is employed with a patience of  5 epochs, depending on the respective phase of training. The model’s performance is assessed based on the Matthews correlation coefficient (MCC), and the best-performing model is retained. Training is conducted over 30 epochs to ensure the generation of robust sequence representations that facilitate accurate protein-protein interaction prediction.

\subsubsection{The contrastive loss coefs ($\kappa$) on model}
To evaluate the impact of the contrastive loss coefficient ($\kappa$) on model performance, we conducted a grid search on the hyperparameter $\kappa$ in equation (12) across four benchmark datasets (Yeast, Human, H. pylori, and PIPR-cut), with values set as {0, 0.3, 0.6, 1.0}. The experimental results, shown in Tables \ref{tab:clc}, indicate significant differences in various performance metrics  for different settings of $\kappa$. The results demonstrate that the appropriate introduction of contrastive loss (such as $\kappa = 0.6$ or $\kappa = 1.0$) can effectively enhance the model’s generalization ability and robustness, validating the effectiveness of the supervised contrastive learning strategy in multimodal feature fusion.

\begin{table}[t]
\centering
\caption{The impact of contrastive loss coefs ($\kappa$) on model output}
\vspace{0.7em}
\label{tab:clc}
\begin{tabular}{l l l l l l l l l l}
\toprule
 Dataset & $\kappa$ & Acc  & Pre  & Recall & Spe  & F1  & MCC  & AUC  & AUPRC \\
\midrule
\multirow{4}{*}{Yeast} 
& 0.0  & 0.9802 & 0.9783 & 0.9821 & 0.9782 & 0.9802 & 0.9603 & 0.9954 & 0.9953 \\
& 0.3 & 0.9808 & 0.9770 & 0.9848 & 0.9768 & 0.9809 & 0.9616 & 0.9967 & 0.9955 \\
& 0.6 & 0.9798& 0.9788& 0.9809& 0.9787& 0.9798& 0.9596& 0.9956& 0.9948\\
& \textbf{1.0} & \textbf{0.9813}& \textbf{0.775}& \textbf{0.9853}& \textbf{0.9772}& \textbf{0.9814}& \textbf{0.9627}& \textbf{0.9969}& \textbf{0.9969}\\
\midrule
\multirow{4}{*}{Human} 
& 0.0  & 0.8285 & 0.8347 & 0.8205 & 0.8366 & 0.8272 & 0.6577 & 0.8983 & 0.9010 \\
& 0.3  & 0.8339 & 0.8464 & 0.8169 & 0.8510 & 0.8310 & 0.6688 & 0.9068 & 0.9091 \\
& 0.6  & 0.8320 & 0.8293 & 0.8366 & 0.8274 & 0.8328 & 0.6643 & 0.9053 & 0.9083 \\
& \textbf{1.0}  & \textbf{0.8302} & \textbf{0.8184} & \textbf{0.8495} & \textbf{0.8110} & \textbf{0.8334} & \textbf{0.6614} & \textbf{0.9058} & \textbf{0.9084} \\
\midrule
\multirow{4}{*}{H.pylori} 
& 0.0   & 0.8721 & 0.8382 & 0.9293 & 0.8134 & 0.8806 & 0.7502 & 0.9382 & 0.9323 \\
& 0.3  & 0.8728 & 0.8381 & 0.9286 & 0.8155 & 0.8809 & 0.7502 & 0.9404 & 0.9327 \\
& 0.6  & 0.8728 & 0.8391 & 0.9273 & 0.8169 & 0.8809 & 0.7498 & 0.9342 & 0.9258 \\
& \textbf{1.0}  & \textbf{0.8767} & \textbf{0.8412} & \textbf{0.9328} & \textbf{0.8190} & \textbf{0.8845} & \textbf{0.7578} & \textbf{0.9399} & \textbf{0.9313} \\
\midrule
\multirow{4}{*}{PIPR-cut} 
& 0.0  & 0.8834 & 0.8769 & 0.8928 & 0.8741 & 0.8846 & 0.7673 & 0.9394 & 0.9192 \\
& 0.3  & 0.8857 & 0.8719 & 0.9048 & 0.8665 & 0.8879 & 0.7722 & 0.9427 & 0.9204 \\
& \textbf{0.6}  & \textbf{0.8854} & \textbf{0.8812} & \textbf{0.8912} & \textbf{0.8797} & \textbf{0.8861} & \textbf{0.7711} & \textbf{0.9446} & \textbf{0.9242} \\
& 1.0  & 0.8836 & 0.8700 & 0.9019 & 0.8652 & 0.8857 & 0.7677 & 0.9407 & 0.9191 \\
\bottomrule
\end{tabular}
\end{table}

\subsubsection{The k-Spaced (k) and similarity threshold ($\tau$) on model}
\begin{figure}[h!]
\centering
\begin{minipage}[h]{0.5\textwidth}
    \centering
    \includegraphics[width=\textwidth]{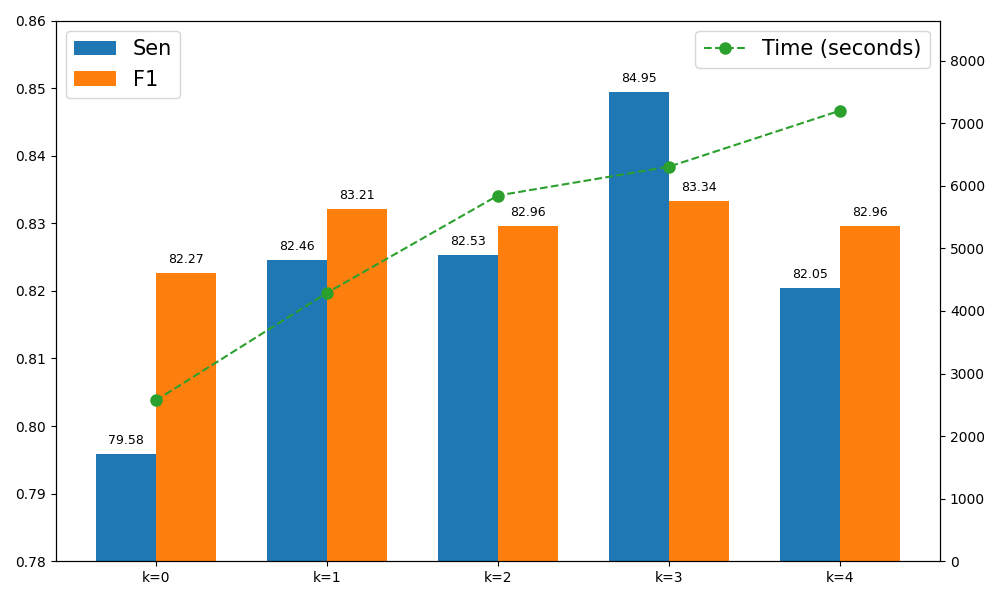}  
    \caption{The impact of k on SCMPPI}
    \label{fig:k}
\end{minipage}%
\begin{minipage}[h]{0.5\textwidth}
    \centering
    \includegraphics[width=\textwidth]{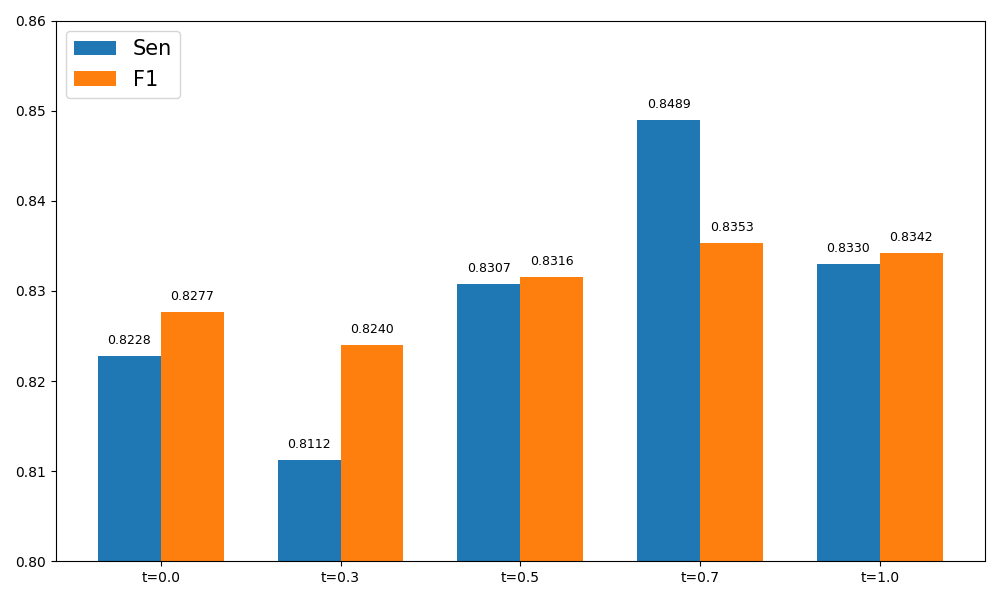}  
    \caption{The impact of $\tau$ on SCMPPI}
    \label{fig:t}
\end{minipage}
\end{figure}

In our experiments using the Human dataset, we optimized two hyperparameters k-Spaced (k) and similarity threshold ($\tau$) in  Figure\ref{fig:k}\ref{fig:t}. 

\begin{figure}[h!]
    \centering
    \includegraphics[width=1\linewidth]{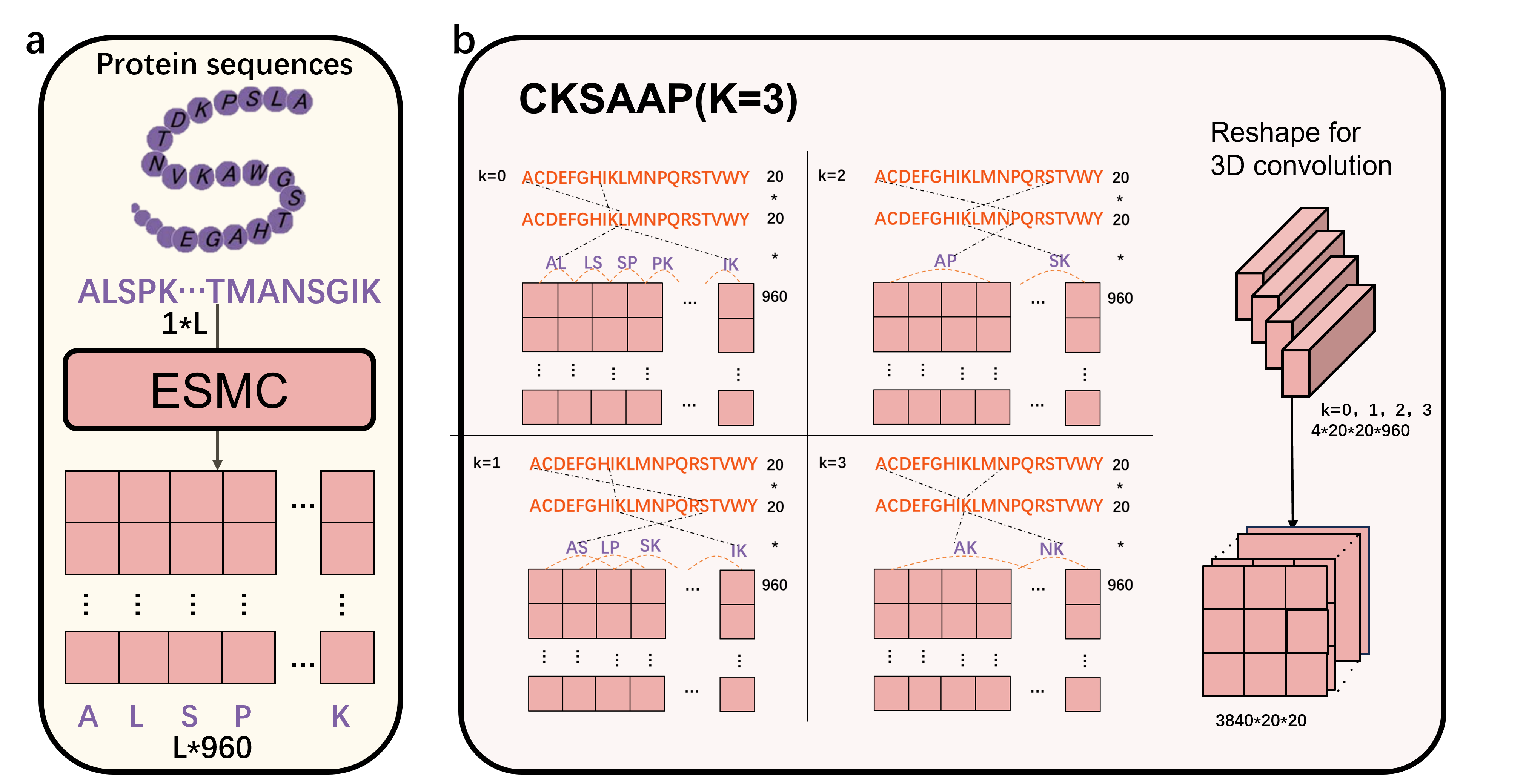}
    \caption{ESMC-CKSAAP module. (a) A protein sequence of length $L$ is passed through a pretrained protein model ESMC to obtain an embedding of size $L \times 960$. (b) The amino acid-level protein embeddings obtained are then mapped into CKSAAP according to a one-to-one correspondence rule, resulting in a new sequence embedding.}
    \label{fig:esmc-cksaap}
\end{figure}

From \autoref{fig:esmc-cksaap}, different k generates embedding representations of different sizes. where the value of k ranges from {0, 1, 2, 3, 4}, and the experimental results show that the model performs best with k=3 after considering the sensitivity, F1 score, and runtime, so we determine k=3 as the final parameter setting (\autoref{eq:k}).

\begin{figure}
    \centering
    \includegraphics[width=1\linewidth]{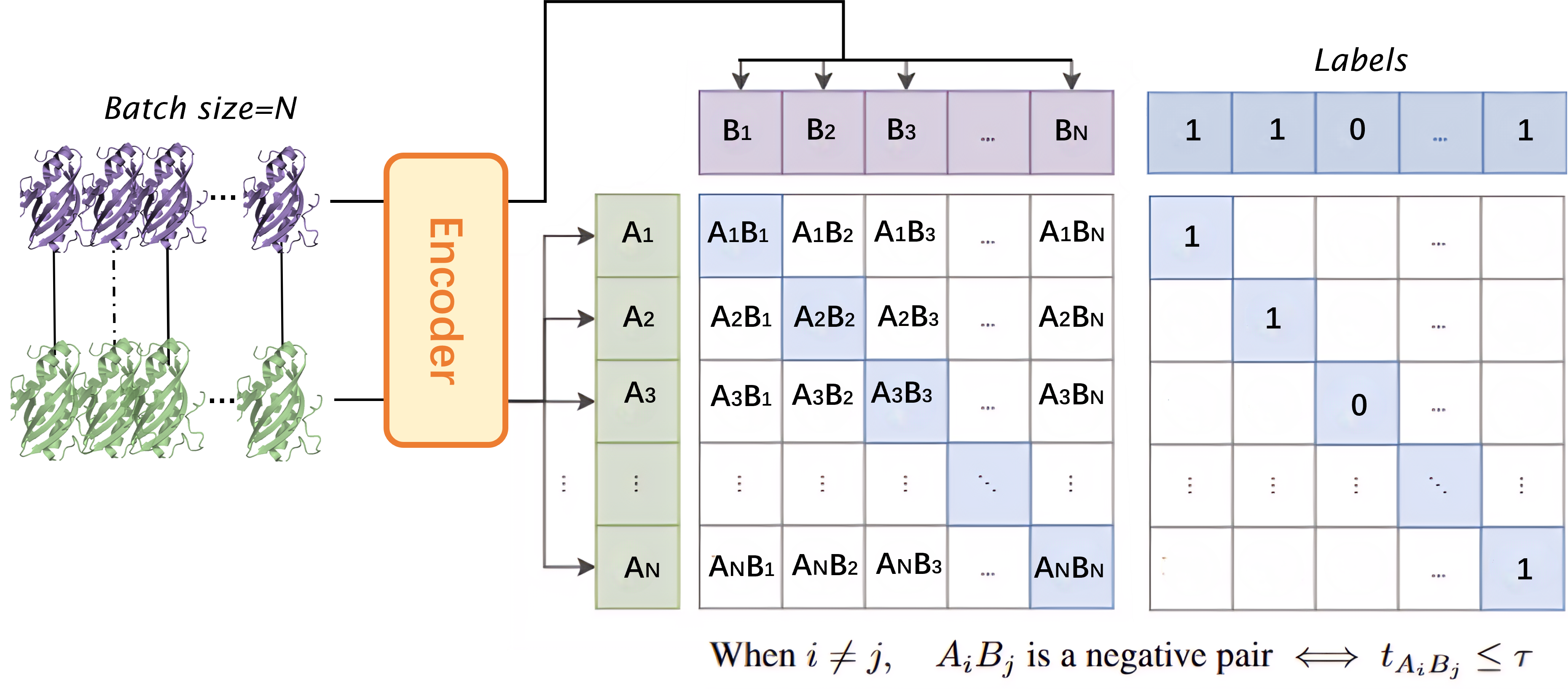}
    \caption{Negative pairs filtering mechanism}
    \label{fig:neg}
\end{figure}
On the other hand, we traversed the range of {0.0, 0.3, 0.5, 0.7, 1.0} to get $\tau$ in \autoref{fig:neg}, and the experimental results showed that the negative sample filter can provide higher sensitivity, lower false negative rate and best F1 score when $\tau$=0.7, so we chose $\tau$=0.7 as the final threshold setting (\autoref{eq:t}) .

\newpage
\section{More results and analysis}
\subsection{More Baseline Results and Ablation}
\begin{table}[h!]
\centering
\caption{Performance comparison on dataset \textit{Human} \textit{($\kappa$=1.0)}}
\vspace{0.7em}  
\label{tab:human}
\resizebox{\textwidth}{!}{
\begin{tabular}{l l l l l l l l}
\toprule
Model & Acc(\%) & Pre(\%) & Sen(\%) & Spe(\%) & F1(\%) & MCC & AUC(\%) \\
\midrule
\textbf{SCMPPI} & \textbf{83.02} & 81.84 & \textbf{84.95} & 81.10 & \textbf{83.34} & \textbf{0.661} & \textbf{90.58} \\
KSGPPI & 81.44 & 82.64 & 79.69 & 83.20 & 81.10 & 0.630 & 85.86 \\
DeepTrio & 75.13 & \textbf{85.90} & 60.51 & \textbf{89.94} & 71.00 & 0.527 & - \\
PIPR & 76.07 & 80.99 & 68.46 & 83.71 & 73.99 & 0.530 & - \\
DeepP-AAC & 72.70 & 73.35 & 72.60 & 72.66 & 72.66 & 0.807 & 80.70 \\
DeepP-CNN & 70.63 & 72.74 & 68.41 & 69.63 & 69.63 & 0.786 & 78.58 \\
DeepFE-PPI & 66.36 & 66.70 & 65.56 & 64.70 & 64.70 & 0.719 & 71.88 \\
\bottomrule
\end{tabular}
}
\end{table}

\begin{table}[h!]
\centering
\caption{Performance comparison on dataset \textit{H.pylori} \textit{($\kappa$=1.0)}}
\vspace{1em}  
\label{tab:pylori}
\resizebox{\textwidth}{!}{
\begin{tabular}{l l l l l l l}
\toprule
  & Acc(\%) & Pre(\%) & Sen(\%) & F1(\%) & AUC(\%) & AP\% \\
\midrule
SCMPPI & \textbf{87.67} & \textbf{84.12} & \textbf{93.28} & \textbf{88.45} & \textbf{93.99} & \textbf{89.65} \\
DFC & 77.14 & 77.11 & 77.17 & 77.09 & 77.18 & 68.87 \\
DCONV & 76.17 & 75.78 & 75.41 & 75.43 & 76.28 & 68.87 \\
DeepP-AAC & 66.14 & 68.10 & 65.62 & 65.67 & 72.86 & 69.31 \\
DeepP-CNN & 64.60 & 66.69 & 63.13 & 63.82 & 71.27 & 69.42 \\
DeepFE-PPI & 61.64 & 61.51 & 62.65 & 61.89 & 66.04 & 63.10 \\
\bottomrule
\end{tabular}
}
\end{table}

\begin{table}[h!]
\centering
\caption{Performance comparison on dataset \textit{pipr-cut} \textit{($\kappa$=0.6)}}
\vspace{1em}  
\label{tab:pipr}
\resizebox{\textwidth}{!}{
\begin{tabular}{l l l l l l l l}
\toprule
  & Acc(\%)  & Pre(\%)  & Sen(\%)  & Spe(\%)  & F1(\%)  & MCC  & AUC(\%)  \\
\midrule
Our Model  & \textbf{88.54} & \textbf{87.49} & 89.12 & \textbf{87.96} & \textbf{88.61} & \textbf{0.771} & \textbf{94.45} \\
KSGPPI  & 88.37 & 87.40 & \textbf{89.70} & 87.05 & 88.53 & 0.768 & 89.96 \\
TAGPPI  & 87.95 & 87.12 & 89.09 & 86.81 & 88.09 & 0.759 & - \\
PIPR  & 86.37 & 89.04 & 83.16 & 84.23 & 85.90 & 0.731 & - \\
DeepTrio  & 84.96 & 85.98 & 83.28 & 86.61 & 84.61 & 0.699 & - \\
DeepFE-PPI  & 74.44 & 73.85 & 75.93 & 72.94 & 74.78 & 0.490 & - \\
\bottomrule
\end{tabular}
}
\end{table}

\begin{table}[h!]
\centering
\caption{Performance comparison on multi-species(\textit{C. eleg, D. mela}
and \textit{E. coli}) dataset ($\kappa$=0.3).}
\vspace{0.7em}  
\label{tab:multi-species}
\resizebox{\textwidth}{!}{
\begin{tabular}{l l l l l l l l l}
\toprule
Model & Acc(\%) & Pre(\%) & Sen(\%) & Spe(\%) & F1(\%) & MCC & AUC(\%) & AP(\%) \\
\midrule
Our Model & \textbf{99.31} & \textbf{99.84} & 98.77 & \textbf{99.84} & \textbf{99.30} & \textbf{0.986} & \textbf{99.84} & \textbf{99.88} \\
HNSPPI & 98.57 & 98.30 & \textbf{98.85} & 94.94 & 98.57 & 0.949 & 98.57 & 92.42 \\
TAGPPI & 99.15 & 99.83 & 98.48 & 99.83 & 99.15 & 0.983 & - & - \\
PIPR & 98.19 & - & - & - & 98.17 & - & - & - \\
DeepP-AAC & 85.14 & 84.40 & 86.65 & - & 85.48 & 0.807 & 80.70 & 81.54 \\
DeepP-CNN & 81.20 & 80.79 & 82.65 & - & 81.59 & 0.786 & 78.58 & 79.66 \\
\bottomrule
\end{tabular}
}
\end{table}

\begin{table}[h]
\centering
\caption{Ablation Study on SCMPPI with three main modules}
\vspace{0.7em}  
\label{tab:xiaorong_comparison}
\resizebox{\textwidth}{!}{
\begin{tabular}{l l l l l l l l l}
\toprule
Dataset & Seq & Graph & Cl & Acc\% & MCC & Sen\% & AUC\% & AUPRC\% \\
\midrule
\multirow{4}{*}{Yeast(0.6)} 
& \checkmark & \checkmark & \checkmark & 98.01 & \textbf{0.960} & \textbf{98.30} & \textbf{99.62} & \textbf{99.68} \\
& \checkmark & \checkmark & \texttimes & \textbf{98.02} & 0.960 & 98.21 & 99.54 & 99.53 \\
& \checkmark & \texttimes & \checkmark & 97.55 & 0.951 & 98.00 & 99.60 & 99.65 \\
& \texttimes & \checkmark & \checkmark & 94.44 & 0.889 & 92.62 & 98.10 & 98.49 \\
\midrule
\multirow{4}{*}{Human(1.0)} 
& \checkmark & \checkmark & \checkmark & \textbf{83.39} & \textbf{0.669} & 81.69 & \textbf{90.68} & \textbf{90.91} \\
& \checkmark & \checkmark & \texttimes & 82.85 & 0.658 & \textbf{83.66} & 89.83 & 90.10 \\
& \checkmark & \texttimes & \checkmark & 82.19 & 0.644 & 81.25 & 89.33 & 89.80 \\
& \texttimes & \checkmark & \checkmark & 81.98 & 0.640 & 82.35 & 88.85 & 89.64 \\
\midrule
\multirow{4}{*}{H.pylori(1.0)} 
& \checkmark & \checkmark & \checkmark & \textbf{87.67} & \textbf{0.758} & \textbf{93.28} & \textbf{93.99} & 93.13 \\
& \checkmark & \checkmark & \texttimes & 87.21 & 0.750 & 92.93 & 93.82 & \textbf{93.23} \\
& \checkmark & \texttimes & \checkmark & 86.97 & 0.743 & 91.56 & 93.32 & 92.90 \\
& \texttimes & \checkmark & \checkmark & 79.74 & 0.596 & 81.62 & 86.82 & 86.10 \\
\midrule
\multirow{4}{*}{PIPR-cut(0.6)} 
& \checkmark & \checkmark & \checkmark & \textbf{88.54} & \textbf{0.771} & \textbf{89.12} & \textbf{94.46} & \textbf{92.42} \\
& \checkmark & \checkmark & \texttimes & 88.34 & 0.767 & 89.28 & 93.94 & 91.92 \\
& \checkmark & \texttimes & \checkmark & 87.88 & 0.758 & 87.09 & 93.84 & 91.79 \\
& \texttimes & \checkmark & \checkmark & 83.43 & 0.669 & 82.88 & 90.10 & 89.33 \\
\bottomrule
\end{tabular}
}
\end{table}

\subsection{Visualization}
\begin{figure}[H]
    \centering
    \includegraphics[width=1\linewidth]{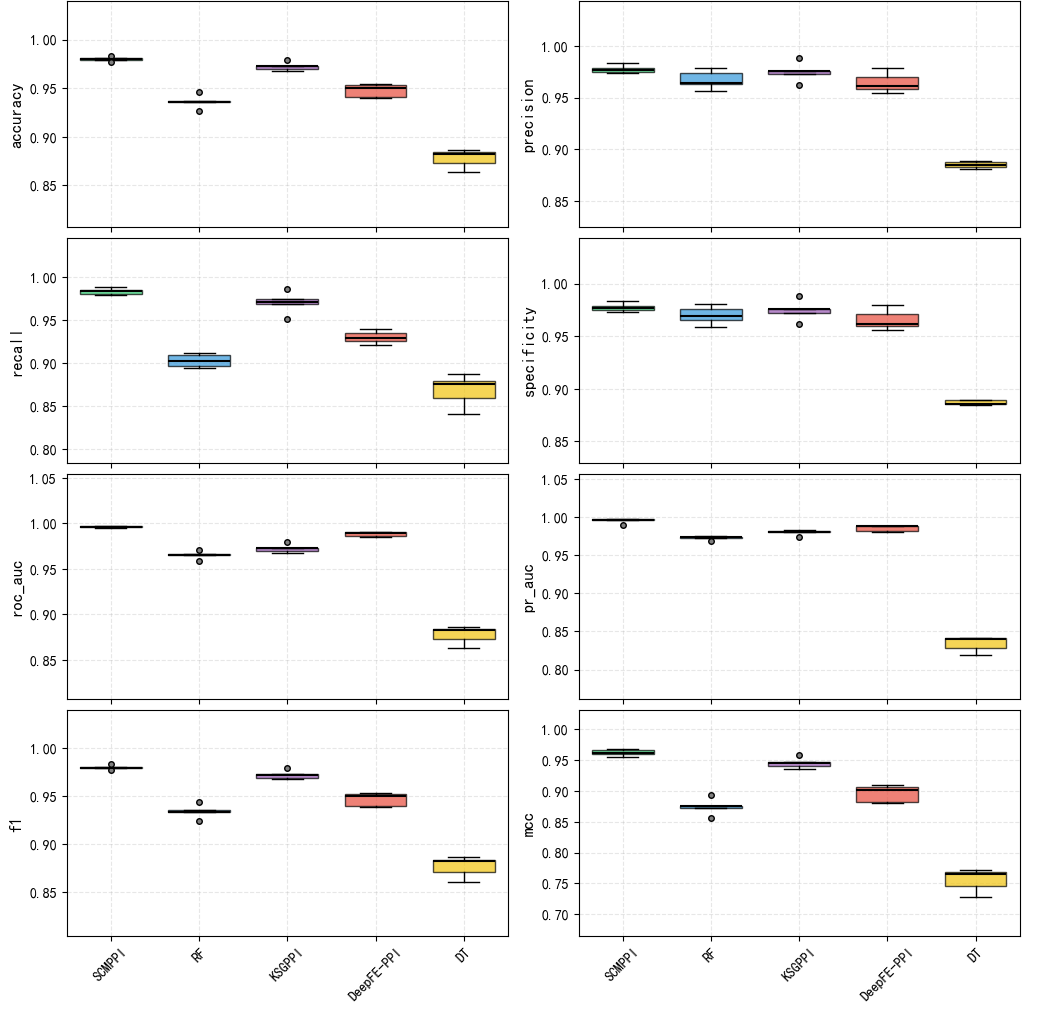}
    \caption{The performance of models through 5-fold cross-validation on the Yeast dataset.}
    \label{fig:box}
\end{figure}

\begin{figure}[ht]
\centering
\begin{minipage}[b]{0.45\textwidth}
    \centering
    \includegraphics[width=\textwidth]{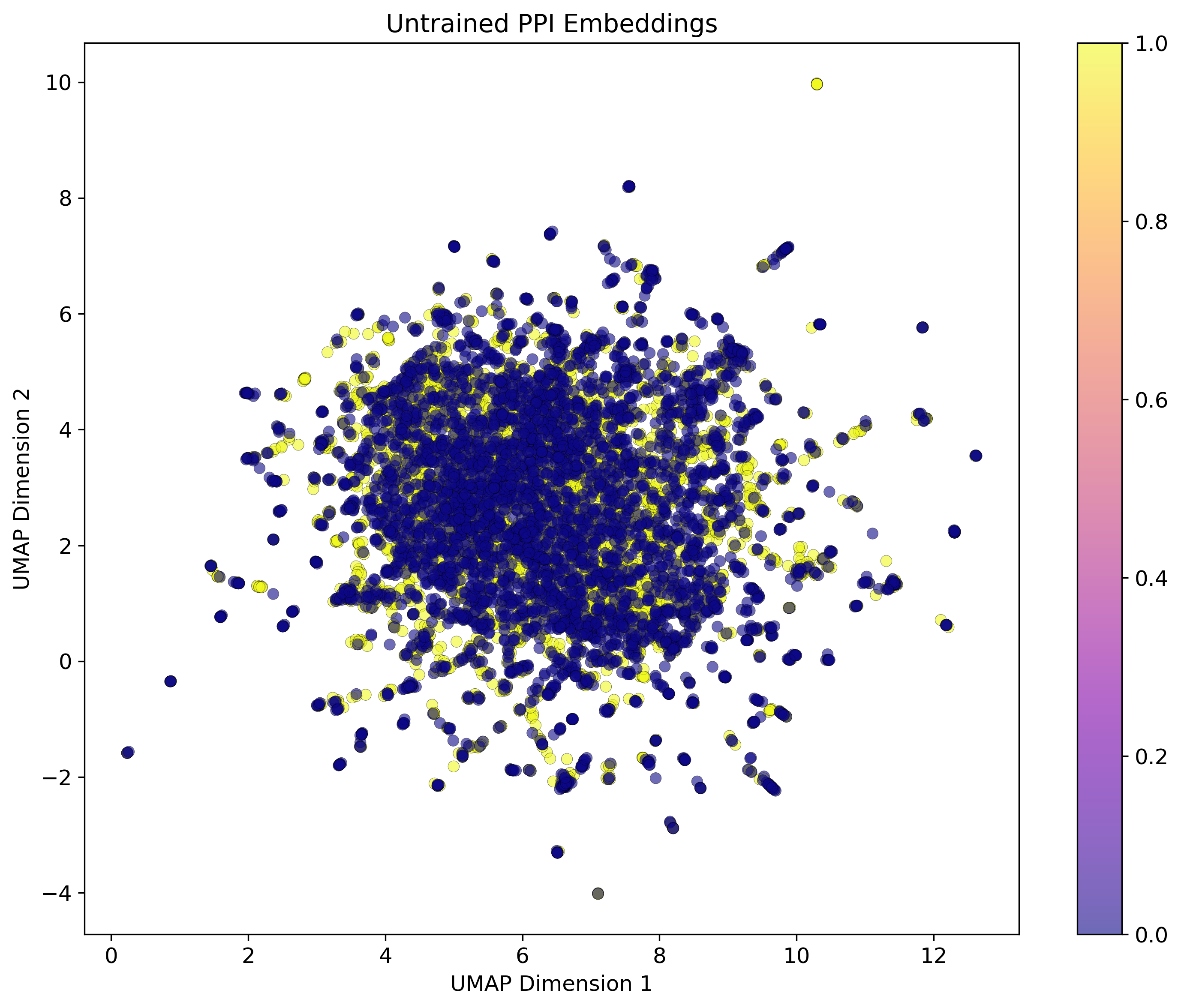}  
    \label{fig:Untrained}
\end{minipage}%
\hspace{0.05\textwidth} 
\begin{minipage}[b]{0.45\textwidth}
    \centering
    \includegraphics[width=\textwidth]{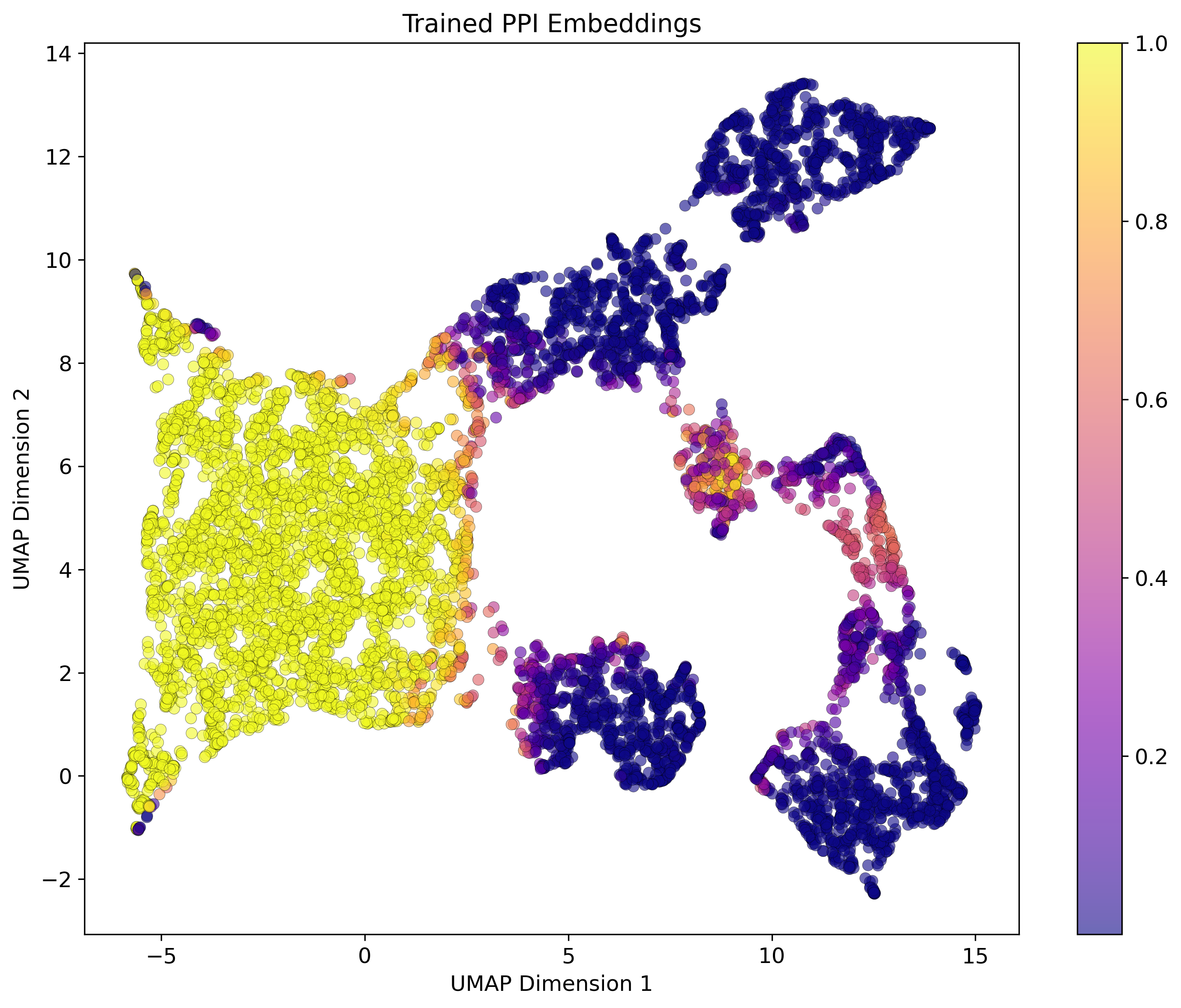}  
    \label{fig:Trained}
\end{minipage}
\caption{PPI classification of Yeast in the UMAP space}
\label{fig:twosubplots}
\end{figure}

\begin{figure}
    \centering
    \includegraphics[width=1\linewidth]{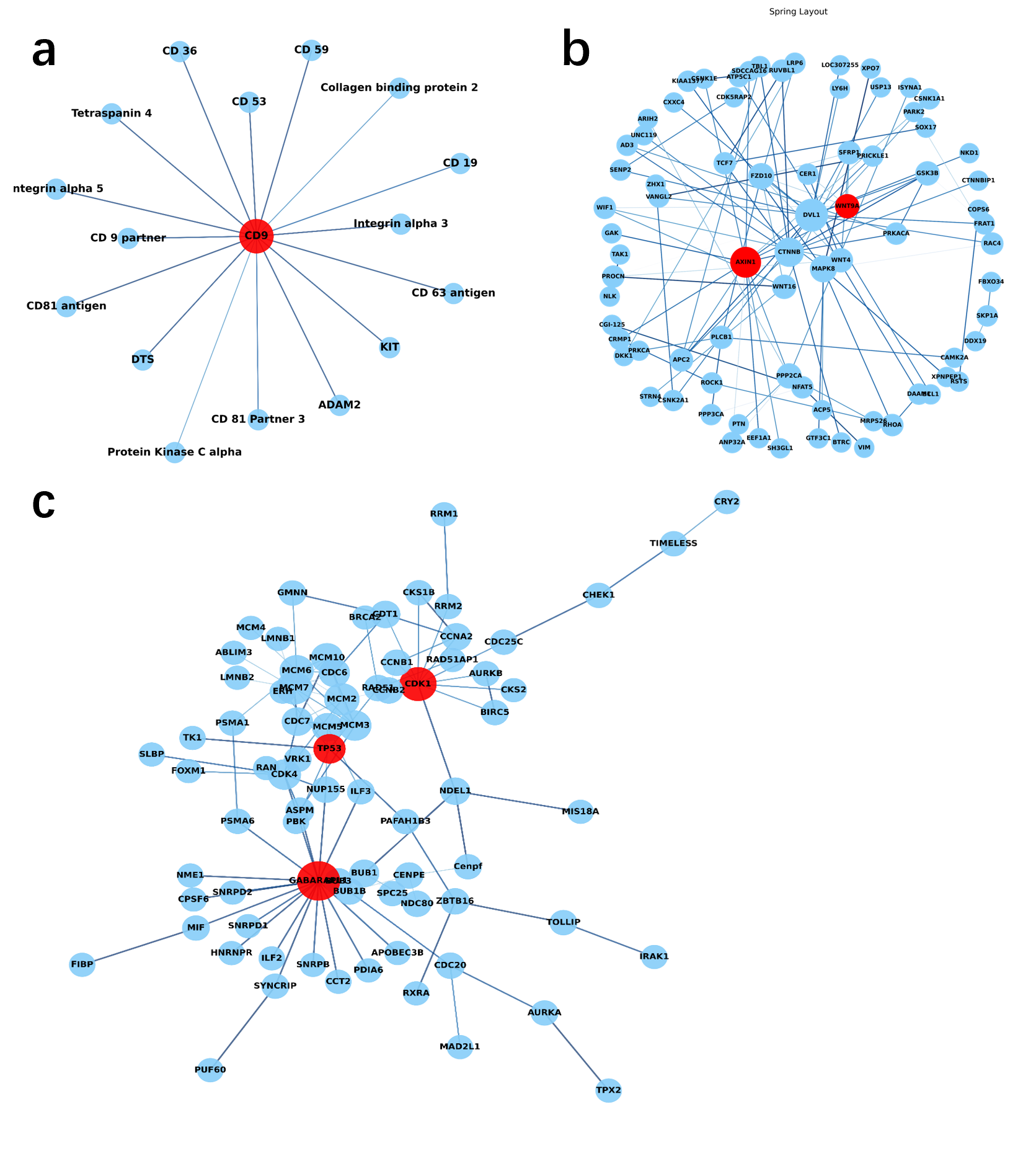}
    \caption{SCMPPI model to predict protein-protein interactions (PPIs) in the CD9(a), Wnt-related pathway(b), and Cancer-specific networks(c)}
    \label{fig:net}
\end{figure}

\end{document}